# Automatically measuring speech fluency in people with aphasia: first achievements using read-speech data

Lionel Fontan [a*], Typhanie Prince [b,c], Aleksandra Nowakowska [b],

Halima Sahraoui [c], and Silvia Martinez-Ferreiro [d]

[a] Archean LABS, Montauban, France

[b] Praxiling UMR 5267, Université Paul Valéry Montpellier 3, Montpellier, France

[c] Neuropsycholinguistics Laboratory (E.A. 4156), University of Toulouse, Toulouse, France

[d] Gerontology & Geriatrics research group, Department of Physiotherapy, Medicine & Biomedical Sciences, University of A Coruña, A Coruña, Spain

[*]Author to whom correspondence should be addressed: Lionel Fontan, Archean LABS, 20, place Prax-Paris, 82000 Montauban, France. Email: lfontan@archean.tech

# Abstract

**Background.** Speech and language pathologists (SLPs) often rely on judgements of speech fluency for diagnosing or monitoring patients with aphasia. However, such subjective methods have been criticised for their lack of reliability and their clinical cost in terms of time.

**Aims.** This study aims at assessing the relevance of a signal-processing algorithm, initially developed in the field of language acquisition, for the automatic measurement of speech fluency in people with aphasia (PWA).

**Methods & Procedures.** Twenty-nine PWA and five control participants were recruited via non-profit organizations and SLP networks. All participants were recorded while reading out loud a set of sentences taken from the French version of the Boston Diagnostic Aphasia Examination. Three trained SLPs assessed the fluency of each sentence on a five-point qualitative scale. A forward-backward divergence segmentation and a clustering algorithm were used to compute, for each sentence, four automatic predictors of speech fluency: pseudo-syllable rate, speech ratio, rate of silent breaks, and standard deviation of pseudo-syllable length. The four predictors were finally combined into multivariate regression models (a multiple linear regression — MLR, and two non-linear models) to predict the average SLP ratings of speech fluency, using a leave-one-speaker-out validation scheme.

**Outcomes & Results.** All models achieved accurate predictions of speech fluency ratings, with average root-mean-square errors as low as 0.5. The MLR yielded a correlation coefficient of 0.87 with reference ratings at the sentence level, and of 0.93 when aggregating the data for each participant. The inclusion of an additional predictor sensitive to repetitions improved further the predictions with a correlation coefficient of 0.91 at the sentence level, and of 0.96 at the participant level.

**Conclusions.** The algorithms used in this study can constitute a cost-effective and reliable tool for the assessment of the speech fluency of patients with aphasia in read-aloud tasks. Perspectives for the assessment of spontaneous speech are discussed.

**Keywords**: speech fluency, automatic assessment, aphasia

**Introduction**

Fluency can be defined as the ease and speed with which a given task is executed. However, when dealing with speech production in people with aphasia (PWA), the notion of fluency becomes far more complex. The first reason is that the fluency of speech production in general can be apprehended from different perspectives (Lickley, 2015): the perspective of the speaker's brain defining a linguistic message and programming its articulation (perspective referred to as *planning fluency*), that of the speech signal produced consequently (*surface fluency*), and that of the impression made on the listener (*perceived fluency*). A second source of complexity arises from the classical clinical distinction between *fluent* and *nonfluent* aphasias. With the exception of conduction aphasia, fluent aphasias, which are generally associated with lesions affecting the posterior part of the central sulcus (Hillis, 2023), are characterised by an ease of articulation and the ability to produce complex sentences, while potentially struggling to access words and making lexical and phonological errors. Nonfluent aphasias, which are traditionally associated with regions anterior to the central sulcus, are in contrast characterised by an effortful speech production, with a reduced ability to articulate, as well as to produce long sentences (Goodglass & Kaplan, 1983; Goodglass et al., 2001; Tarulli, 2021). From a linguistic point of view, the criteria used to distinguish between the two groups of aphasias are thus complex, involving different levels of analysis (i.e., phonetic, phonological, lexical and syntactic). One particularly confusing point is that, according to these criteria, even "fluent" aphasias can cause disfluencies at the surface level, for example with the increase of silent pauses due to the speakers struggling while searching for particular words, that will in turn impact perceived fluency (Andreetta & Marini, 2015; Edwards, 2005).

These sources of complexity and ambiguity may, at least partially, explain why a poor reliability is often observed in subjective judgements of the fluency of PWA (Clough & Gordon, 2020; Gordon, 1998; Gordon & Clough, 2022; Kerschensteiner et al., 1972; Poeck, 1989). These measures are generally used to categorise/diagnose PWA according to the fluent-nonfluent dichotomy, or as continuous variables that can serve as quantitative indicators for characterizing speech impairments (Clough & Gordon, 2020). From a clinical point of view, the lack of reliability is a highly critical issue since today the majority of speech-language pathologists (SLPs) working with PWA use subjective methods to assess the fluency of their patients (Gordon & Clough, 2022).

As a consequence, there is a crucial need for the development of new, more reliable ways of measuring fluency that could be used for clinical purposes with PWA. As underlined by Clough and Gordon (Clough & Gordon, 2020; Gordon & Clough, 2022), some desirable features of such new fluency metrics would be being continuous rather than categorical, multidimensional rather than bimodal, and objective rather than subjective. The continuity and multimodality of the metrics are meant to overcome the limitations of the too simplistic fluent/nonfluent binary categorization, that has been criticised for long (Poeck, 1989) and leave room for the definition of multidimensional "fluency profiles". Those profiles would account for the fluency of PWA at different levels of analysis — for example, at one or several of the three levels of increasing linguistic complexity proposed by Clough and Gordon (2020): speech production, lexical retrieval, and grammatical competence. The objective nature of the metrics would guarantee their reliability; however, as some objective methods can be very time-consuming (e.g., a manual measurement of speech rate), a last desirable feature for the new methods would be being easily practicable in the clinical context (Gordon & Clough, 2022).

To cope with the lack of reliability and objectivity of fluency judgements, a recent attempt was made by Metu et al. (2023) to use machine-learning algorithms for predicting speech fluency in PWA, as evaluated by SLPs on the Western Aphasia Battery-Revised (WAB-R; Kertesz, 2006). They used neural networks, taking as input either images (spectrograms), or Mel-frequency cepstral coefficients (MFCCs) and I-vectors extracted from audio recordings. However, their results showed that the judgements of the SLPs on the WAB-R scale lacked reliability in terms of inter-rater agreement; also, their machine-learning algorithms were not as useful for predicting the WAB-R fluency ratings as "simple" trichotomous judgements made by SLPs who were asked to categorise each PWA as fluent, nonfluent or mixed.

The problem of time-consuming, potentially unreliable subjective evaluations of fluency is also encountered in the domain of first-language (L1) or second-language (L2) acquisition, in which teachers and researchers often need to assess and monitor learners' speech production skills (Cucchiarini et al., 2000; Detey, et al., 2020; Fontan et al., 2022). More precisely, acquisition studies focus on the assessment of speech fluency at the phonetic, surface level, that is, on the extent to which speech produced by L1/L2 speakers "flows easily without pauses and other disfluency markers" (Derwing & Munro, 2015). Such markers include a number of disfluencies that are also taken into account, along with speech rate and pauses, in the field of aphasiology for characterizing speech fluency: repetitions, repairs and false starts (Faroqi-Shah et al., 2020; Gordon & Clough, 2022; Jacks & Haley, 2015; Wang et al., 2013); they do not, however, include disfluencies occurring at higher linguistic levels, that is, at the lexical and grammatical levels. The studies in L1/L2 acquisition therefore focus on the "speech production" level of speech fluency (Clough & Gordon, 2020) — which is also the scope of the present study.

To develop objective and rapid measures of speech fluency, the earliest works in L2 acquisition proposed using metrics based on automatic speech recognition (ASR), such as automatically computed estimates of speech and articulation rates (Cucchiarini et al., 2000; 2002). One problem with ASR systems is that their results can also lack reliability when processing "atypical" speech such as nonnative speakers' or children's speech (Gelin et al., 2021; Vu et al., 2014), as well as disordered speech (De Russis & Corno, 2019) — including that produced by PWA (Jamal et al., 2017), which could in turn bias metrics of speech fluency (e.g., by leading to an under- or overestimation of speech rate if shorter or longer words as those actually pronounced are recognised by the ASR system). More recent studies have thus proposed discarding ASR and using instead automatic, direct acoustic measures for computing predictors of speech fluency. Fontan et al. (2018) first used the forward-backward divergence segmentation (FBDS) algorithm (André-Obrecht, 1988) to measure speech fluency in Japanese learners of French. Contrary to ASR systems, the FBDS algorithm performs an analysis of speech signals in the temporal domain only. More precisely, the algorithm focuses on the trajectory of the energy over time; whenever a significant change is detected (e.g., an abrupt increase of the energy), the corresponding moment is recorded as a segment boundary. This process leads to a segmentation in sub-phonemic units that have been shown to correspond to different stages involved in the articulatory production of speech segments (André-Obrecht, 1988).

Fontan et al. (2018) used FBDS segments to compute predictors of speech fluency, such as estimates of speech rate and of the frequency of occurrence of filled and silent pauses. When combined in a multivariate model, these estimates could accurately predict speech fluency, as rated by native speakers on a qualitative scale. The proof-of-concept was later extended to several other L2 and L1 healthy populations (Fontan et al., 2019, Fontan et al., 2020, Kondo et al., 2020, Fontan et al., 2022), and refinements in the methods led to prediction models explaining up to 94% of the variance in perceived speech fluency (Fontan et al., 2022). Among these refinements, the most recent studies included an automatic clustering of FBDS segments into silent breaks and pseudo-syllables. As opposed to "true" syllables whose boundaries are identified based on human perception and/or manual acoustic analyses, Farinas and Pellegrino (2001) and Rouas et al. (2005) used the term of *pseudo-syllables* to refer to automatically identified clusters of FBDS speech segments consisting a vowel and possibly of one or several consonants. Fontan et al. (2022) showed that adding this clustering step before the computation of predictors significantly improves the prediction of perceived speech fluency.

Such signal-processing algorithms could be of great interest to the field of aphasiology. They provide objective, reliable measures of fluency at the speech production level that could be part of the multidimensional fluency profiles that are called for in the domain. The automatic estimates of speech rate and of the number of silent or filled pauses they rely on are also used for assessing fluency in PWA (Gordon & Clough, 2022; Wang et al., 2013), with some of these indicators also being used for classifying subtypes of aphasia (Ash et al., 2013; Fraser et al., 2014; Potagas et al., 2022). Moreover, as they only process speech signals in the temporal domain, the algorithms require very low computing resources, which make their execution on standard PCs very rapid. They are also meant to be robust to noisy conditions, as the threshold used to distinguish speech from silence is relative (Fontan et al., 2022). The algorithms could therefore be used in suboptimal, real-life clinical scenarios where time is often limited, and high-end audio-recording devices and acoustically-treated rooms might not be available for recording patients.

**Aims**

The main aim of the present study is to investigate whether such signal-processing techniques can be used to predict the speech fluency of PWA, as evaluated by experienced SLPs. More precisely, the study focuses on the prediction of *read-speech* fluency, that is, of sentences read out loud by PWA and control participants. The choice of using, as a first step, speech data collected during a reading task is motivated by the ability to compare the results with those of Fontan et al. (2022 and previous studies) who used similar materials and methods, and also by the fact that the prediction of spontaneous speech fluency is a much more challenging task (Cucchiarini et al., 2000; 2002), and is thus regarded as a longer-term objective.

A secondary aim of the study is to investigate whether using nonlinear models can yield better prediction outcomes than a multiple linear regression. This is motivated by the assumption that some of the predictors, or their relationship, might not be of equal importance all along the speech fluency scale (e.g., the presence of silence breaks might have a lesser impact on the lower end of the scale, when speech rate is very low, than at the higher end). Consequently, as in Fontan et al. (2022), two nonlinear models are used in complement to a multiple linear regression: a support vector regressor with a radial kernel and a random forest regressor.

**Materials and methods**

*Speech recordings*

*Participants*

Thirty-four adult, native-French participants were recruited for the study, among which 29 (eight female) were diagnosed with chronic aphasia, and five (two female) were recruited as control participants. At the time of the study, the age of the participants with aphasia ranged from 27;1 to 74;6 y.o. (years old; *mean*: 60;4; standard deviation, *SD*: 11;9). The age of control participants ranged from 23 to 74;10 y.o. (*mean*: 54;4; *SD*: 22;5).

Both control participants and PWA had self-reported normal, or corrected-to-normal visual and hearing abilities. PWA in the sub-acute or chronic stage were recruited based on confirmed aphasia diagnosis (including Primary Progressive Aphasia), with no history of any other neurological disorder. For PWA the time post-onset of aphasia ranged from 0;8 to 25 years (*mean*: 11;1; *SD*: 16;7).

In order not to put the participants at risk due to the COVID-19 pandemic context that set up during the course of the study, the following non-inclusion criteria were also applied: unstable medical condition, obesity, respiratory failure, and immuno-depressive therapy. Additional individual background information on the participants is provided in appendix A.

All participants were volunteers. PWA were recruited through ads sent to non-profit associations of PWA attached to the French National Federation of PWA (*Fédération Nationale des Aphasiques de France* – FNAF[1]), and to SLP local networks. Before participating in the study, which was approved by the Research & Ethical Committee of Toulouse Federal University (France; file number: 2020-268), all participants provided their informed consent.

---

[1] https://aphasie.fr

*Data collection*

The data collection took place as part of a larger research protocol (Sahraoui et al., 2022) designed for the AADI project[2] (*Aphasie et Discours en Interaction* [Aphasia And Discourse in Interaction]) that aims at creating a database of oral data collected from French-speaking PWA through five speech elicitation tasks: a sentence reading task (see below), a storytelling task (free recall of a story like Cinderella), a picture description task (using the "Cat rescue" image; Nicholas & Brookshire, 1993), an autobiographical interview, and a free oral interaction (conversational dyad).

The present study uses exclusively data from the read-aloud task, in which participants were asked to read aloud a series of 10 sentences of increasing complexity taken from the French version of the Boston Diagnostic Aphasia Examination (Goodglass et al., 2001; Mazaux & Orgogozo, 1982). Due to the COVID-19 pandemic, this task was administered online, using the Zoom platform (Zoom Video communications, San Jose, CA). All participants used their own PCs. Sound was recorded from the PC built-in microphones, using a 16-bit quantization and a 44100 Hz sampling rate.

**Subjective assessment of speech fluency**

*Speech materials*

The recordings corresponding to the three longest sentences (in terms of syllable count) of the read-aloud task materials were used to collect reference speech-fluency ratings.

---

[2] https://www.researchgate.net/project/Aphasia-and-interactive-discourse-analysis-creating-a-database-and-new-methods-for-exploiting-data-AADI

The three sentences consisted of nine words (13 syllables), eight words (11 syllables), and eight words (12 syllables), respectively. The choice of using the longest sentences was motivated by the fact that with longer sentences more disfluencies can potentially occur, which can increase the variability in the speech production performances of PWA and, consequently, the variability in subjective fluency ratings.

Five of the PWA had significant difficulties in producing one or two out of the three target sentences, in which cases the experimenter asked them to move on to the next sentence in order to avoid frustration and discouragement. These difficulties concerned each of the three target sentences, and led to the loss of seven recordings out of the 102 that would have been otherwise expected. As a consequence, the final corpus contained 95 recordings (corresponding to 34 participants × 3 sentences − 7 fails). All recordings were downsampled to 16 kHz, as this sample rate was required by some of the signal-processing algorithms later used to predict speech fluency.

*Rating procedure*

Three SLPs, all of them women, participated in the rating task. None of the SLPs knew any of the PWA recruited for the present study. The SLPs were aged 32, 35, and 52 years old and benefited from eight, 11, and 28 years of professional experience, respectively. All three SLPs were native speakers of French, and did not report any history of hearing difficulty.

Each SLP completed the rating task individually, in a quiet room. A custom-written Python programme was used for the presentation of speech recordings on a MacBook Air 13-inches laptop computer (Apple Inc., Cupertino, CA). The raters listened to the recordings through Audio-Technica ATH-M50x circumaural headphones (Audio-Technica Ltd., Machida, Japan). They were instructed to rate each recording on a five-point scale ranging from 1 (very poor fluency) to 5 (excellent fluency), which was displayed under the form of radio buttons on the computer screen. The raters could listen to each recording as many times as required before rating it.

The 5-point fluency rating scale designed for the present study was not used by any of the SLPs in their professional practice. As a consequence, prior to the rating task, the SLPs were first familiarized with the rating scale by listening to recordings illustrating the whole range of the scale. The recordings used as examples were not part of the speech materials later used in the proper rating task, and were not recorded from any of the PWA recruited for the present study.

### *Automatic extraction of predictors of speech fluency*

*Forward-backward divergence segmentation and clustering*

Each recording was automatically segmented using the FBDS algorithm (André-Obrecht, 1988). When processing a given audio signal, the FBDS algorithm performs two analyses: one forward (i.e., from the beginning of the recording to the end of the recording) and one backwards (i.e., in reverse order). During each of these analyses, the algorithm detects significant changes in the trajectory of the signal energy. Such changes generally correspond to boundaries between different phones or even between different stages of the production of a given phone (e.g., between the onset and the sustain of a vowel), hence resulting in a sub-phonemic segmentation of the input signal. At the end of the process, the boundaries found during the forward and backward analyses are merged together.

FBDS segments were automatically clustered into pseudo-syllables and silent breaks, as this procedure was shown to improve the prediction of speech fluency (Fontan et al., 2022). To this end, FBDS segments were first classified as speech or silent segments as a function of the ratio between their maximum energy and the maximum energy found in the whole recording. Consecutive silent segments were merged together, and the resulting segments were identified as silent breaks if their total duration exceeded 250 ms (De Jong & Bosker, 2013). As the process was fully automatic, such breaks could occur between words as well as within a word. Consecutive speech segments were considered as part of the same pseudo-syllable if the signal energy did not decrease below a certain threshold when switching from the first segment to the next. An example of the results of these automatic segmentation and clustering steps is shown in Figure 1.

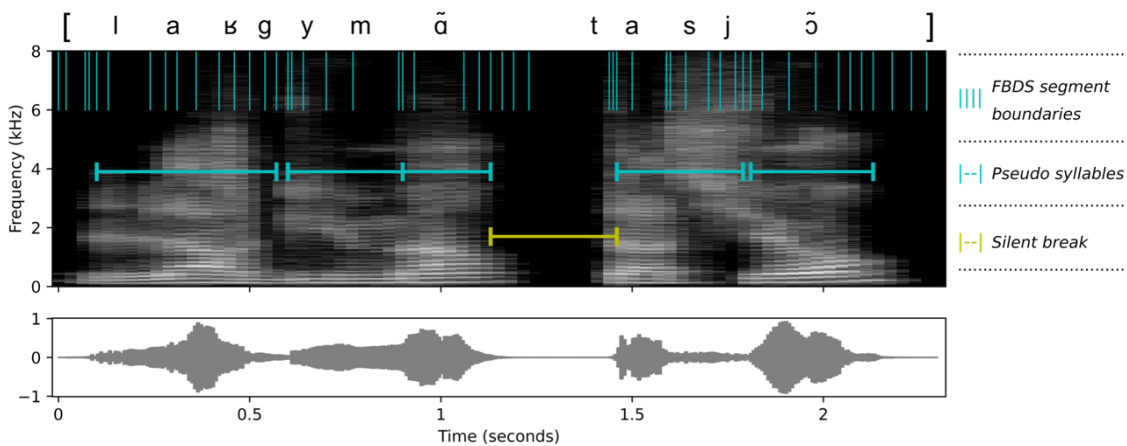

**Figure 1**. Results of the FBDS segmentation and clustering for the word "*l'argumentation*" ([laʁgymɑ̃tasjɔ̃], "the argumentation") produced by one the participants with aphasia, superimposed over the spectrogram of the corresponding audio signal. The IPA-based phonetic transcription appearing above the top panel was manually done by the authors. The bottom panel shows the waveform of the signal.

*Computation of predictors of speech fluency*

For each of the 95 recorded sentences, four predictors of speech fluency were calculated based on pseudo-syllables and silent breaks:

(1) Pseudo-syllable rate, calculated as the number of pseudo-syllables divided by the duration of the recording (in ms). As this predictor aims at estimating speech rate, it is supposed to be positively correlated with ratings of speech fluency.

(2) Standard deviation of pseudo-syllable duration (in ms). This predictor is assumed to be negatively correlated with ratings of speech fluency. Its value should increase with the presence of abnormally long or abnormally short pseudo-syllables, as in the case of filled pauses or false starts, respectively. For example, for the word "surgit" ([syʁʒi], "appears suddenly"), which is part of the linguistic materials used in the present study, short false starts such as [z] or [ʃ] may occur in place of the triphonemic syllable [syʁ].

(3) Speech ratio, calculated as the total duration of pseudo-syllables (in ms) divided by the total duration of the recording (in ms). The value of this estimate should decrease with the presence of silence (i.e., silent breaks and silences shorter than 250 ms). The speech ratio is therefore supposed to be positively correlated with ratings of speech fluency.

(4) Rate of silent breaks, calculated as the number of silent breaks divided by the total duration of the recording (in ms), and assumed to be negatively correlated with ratings of speech fluency.

*Prediction of speech fluency*

Three models were used to predict speech fluency ratings, among which a multiple linear regression (MLR) and two non-linear models: a support vector regression (SVR) using a radial kernel, and a random forest regressor (RFR). The SVR and RFR models were fed with all four predictors. As the accuracy of multivariate linear predictions can be negatively affected by collinearity, the MLR model implemented a stepwise selection, during which any predictor whose contribution was not significant at the 5% level was eliminated.

Given the limited size of the dataset, the three models were trained and evaluated using a leave-one-speaker-out (LOSO) setup. This means that for each model type (MLR, SVR, and RFR), 34 models were actually developed using the data from $n-1$ participants (i.e., from 33 participants) for training purposes, and the data from one single participant (i.e., the speaker out of the training set) for testing purposes. Each of the 34 participants was used as a test participant, and the goodness-of-fit of each model type was finally assessed by computing the average root-mean-square-error (RMSE) between actual and predicted speech-fluency ratings for the total number of participants.

Prior to the evaluation stage, several hyperparameters of the SVR and RFR were tuned: the regularization parameter and epsilon value for the SVR, and the number and maximum depth of the decision trees for the RFR. A nested-cross-validation (LOSO) procedure was used for this fine-tuning, as the creation of a separate validation set was not possible due to the limited size of the dataset.

*Statistical analyses*

All prediction models were built with the Scikit-learn 1.1.2 Python library, and their performances were assessed through custom-written Python scripts using Pandas 1.4.3 and Scipy 1.8.1 libraries.

As far as inferential statistics are concerned, appropriate nonparametric tests were used whenever the prerequisites for parametric tests were violated. The data visualizations provided in the different Figures were generated using Matplotlib 3.5.2, Seaborn 0.12.0., and Librosa 0.9.1 Python libraries.

**Results**

*Reliability of subjective speech-fluency ratings*

*Intra-rater reliability*

As each rater assessed the whole set of stimuli twice, intra-rater reliability could be assessed by checking the consistency of the ratings provided for the same stimuli. For each rater, Spearman rank correlation coefficients (*rhos*) and Cronbach *alphas* were computed between first and second ratings (Table 1).

**Table 1**. Spearman rank correlation coefficients and Cronbach alphas computed between the first and second speech-fluency ratings provided by each rater for the same stimuli.

| Rater | Spearman's *rho* | Cronbach's *alpha* |
|---|---|---|
| 1 | 0.87*** | 0.94 |
| 2 | 0.76*** | 0.88 |
| 3 | 0.80*** | 0.90 |

The high-to-very-high positive correlation coefficients and *alphas* indicate a high intra-rater reliability. As a consequence, the speech-fluency ratings provided by each rater for the same stimuli were averaged for subsequent analyses.

*Inter-rater reliability and agreement*

To check inter-rater reliability, Spearman *rhos* were computed between the ratings provided by each couple of raters (Table 2).

**Table 2**. Spearman rank correlation coefficients computed between speech-fluency ratings provided by the three raters.

|         | Rater 2  | Rater 3  |
| ------- | -------- | -------- |
| Rater 1 | 0.87***  | 0.89***  |
| Rater 2 |          | 0.86***  |

*** *p* < .001 (one-tailed)

The very strong and positive correlation coefficients (all *rhos* ≥ 0.86) indicate an excellent reliability, which is confirmed by a very high *alpha* value of 0.95. A Kruskal-Wallis test indicated that no significant difference existed between the three distributions, $H(2) = 1.8$, $p = 0.41$. Consequently, speech-fluency ratings were averaged across raters to compute the final reference ratings used for the prediction algorithms. Figure 2 shows the distribution of these ratings for participants with and without aphasia.

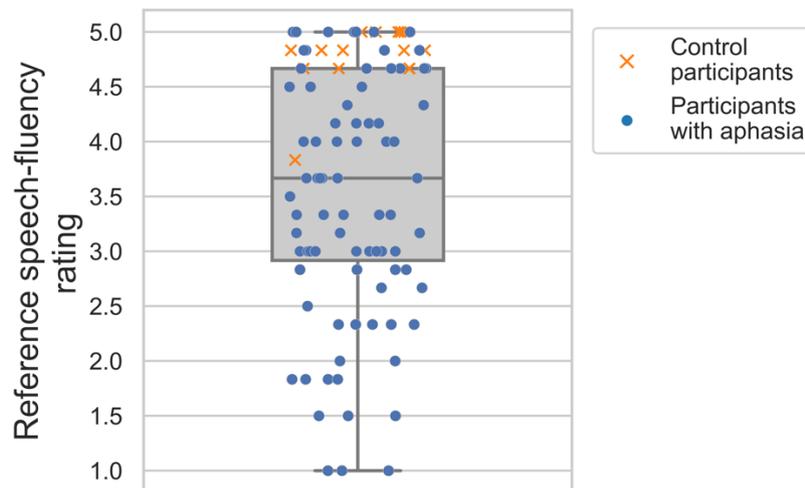

**Figure 2**. Boxplot showing the distribution of reference speech-fluency ratings for PWA and control participants. The horizontal line inside the box corresponds to the median rating. The bottom and top lines of the box represent the 25th and 75th percentiles, while bottom and top whiskers correspond to the minimum and maximum values, respectively.

As can be observed, the final reference ratings span the entire range of the fluency scale, with minimal and maximum values of 1 and 5, which is desirable for regression analysis. As could be expected, the ratings associated with control participants are located in the higher end of the scale, with all values bar one exceeding 4.5.

*Relationship between automatic predictors and speech-fluency ratings*

Table 3 shows Spearman correlation coefficients observed between each automatic predictor of speech fluency and reference ratings. Correlations are strong to very strong, with absolute coefficients ranging from 0.63 (for standard deviation of pseudo-syllable duration) to 0.87 (for pseudo-syllable rate). As was assumed, pseudo-syllable rate and speech ratio are positively correlated with subjective ratings of speech fluency, while standard deviation of pseudo-syllable duration and ratio of silent breaks are negatively correlated with the ratings.

**Table 3**. Spearman rank correlation coefficients computed between each automatic predictor and speech-fluency ratings.

| Predictor | Spearman's *rho* |
|---|---|
| Pseudo-syllable rate | 0.87*** |
| Standard deviation of pseudo-syllable duration | −0.63*** |
| Speech ratio | 0.71*** |
| Ratio of silent breaks | −0.72*** |

\*\*\* *p* < .001 (one-tailed)

**Prediction of speech fluency ratings**

Table 4 presents the average RMSE computed between actual and predicted fluency ratings using the MLR, SVR, and RFR. As can be observed, all three model types yield rather accurate predictions, with average RMSEs equal or below 15% of the speech-fluency scale.

Table 4. Average root-mean-square errors (RMSEs) and standard deviation associated with each of the prediction models.

| Model type | Average RMSE | Standard deviation |
|---|---|---|
| MLR | 0.51 | 0.23 |
| SVR | 0.51 | 0.26 |
| RFR | 0.59 | 0.34 |

A linear mixed model, using the average RMSE observed for each participant as the dependent variable, the speaker-out as a random effect, and model type as a fixed effect, confirmed that no significant difference existed between the RMSEs achieved by the three model types, $F(2, 99) = 0.9$, $p = 0.41$. As a consequence, the MLR was used for subsequent analyses, as it is the simplest and most interpretable of the three models.

Figure 3 compares reference subjective fluency ratings to those predicted using the MLR. The MLR achieved a very high correlation with reference ratings, with a correlation coefficient of 0.87 (see left panel of Figure 3). As, in clinical practice, the speech fluency of a given patient would likely be evaluated using more than one sentence, reference and predicted ratings were also aggregated for each participant (i.e., up to three sentences were considered for each participant). The resulting correlation is stronger, with a Pearson's coefficient of 0.93. The accuracy of the predictions is also higher, with a RMSE of 0.4, corresponding to one tenth of the speech fluency scale (see the right panel of Figure 3).

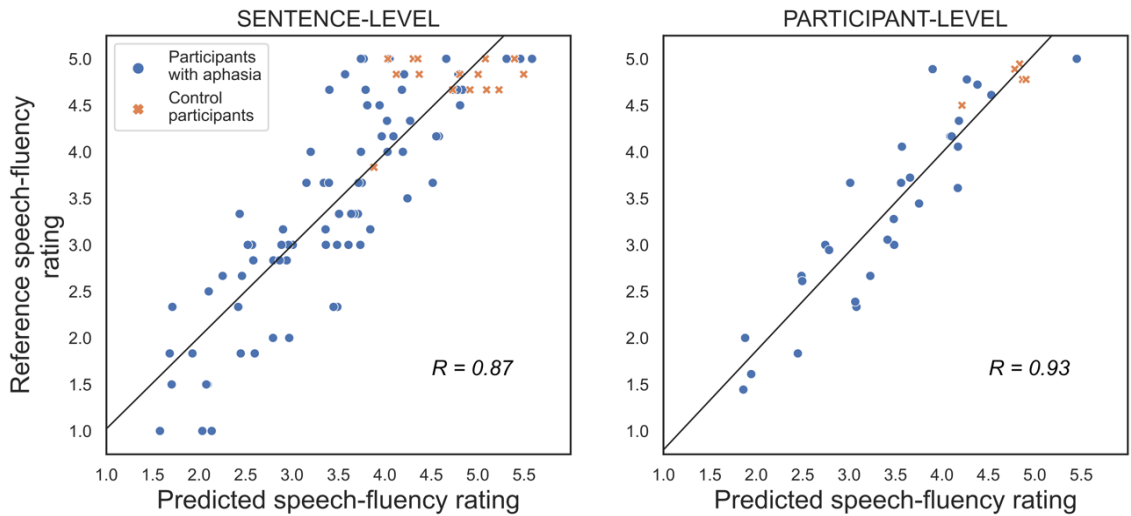

**Figure 3**. Scatterplots relating reference speech-fluency ratings to ratings predicted using the MLR, and associated regression lines. The left panel shows data at the sentence level (95 data points), while the right panel shows data averaged for each participant (34 data points).

To assess the importance of each automatic predictor, an additional MLR was computed to predict the whole 95 reference speech-fluency ratings. According to the resulting coefficient of determination, the model explained 78.1% of the variance of speech-fluency ratings, based on three predictors: speech ratio, pseudo-syllable rate, and standard deviation of pseudo-syllable rate. The rate of silent breaks was eliminated during the stepwise selection, probably due to its very strong correlation with speech ratio, Spearman's *rho* = −0.90, $p < 0.001$. Table 5 presents the three standardised (*beta*) coefficients associated with each automatic predictor, by order of importance. Speech ratio is the automatic measure that contributes the most to the prediction of speech fluency. Pseudo-syllable rate and standard deviation of pseudo-syllable duration are of similar importance to the model.

**Table 5**. Standardised (*Beta*) coefficients associated with each automatic predictor when computing a multiple linear regression.

| Predictor | *Beta*-coefficient |
|---|---|
| Speech ratio | 0.47 |
| Pseudo-syllable rate | 0.31 |
| Standard deviation of pseudo-syllable duration | −0.30 |

**Discussion**

In this study, the signal-processing algorithms used by Fontan et al. (2022) for measuring speech fluency in L1 and L2 healthy speakers were used to predict speech fluency ratings given by three trained SLPs for PWA and control participants. As, in aphasiology, subjective judgements of speech fluency have been criticised for their lack of reliability (Clough & Gordon, 2020; Gordon, 1998; Gordon & Clough, 2022; Kerschensteiner et al., 1972; Poeck, 1989), a first task was to check the consistency of SLP ratings. Very high intra- and inter-rater agreements, at the same level as those observed in previous studies in language acquisition using similar fluency scales, were obtained. For example, the minimum (0.86) and mean (0.87) inter-rater correlation coefficients obtained in this study were exactly the same as those in Fontan et al. (2022). These high agreement rates might be explained, at least partially, by the fact that in the present study fluency was judged as a continuous variable rather than using the clear-cut fluent/non fluent dichotomy potentially more prone to disagreements (Gordon & Clough, 2022). Moreover, the current study focused on read speech, which is certainly less complex to assess than spontaneous speech, for which no reference as to the intended message is available to the raters, and that may involve speech production errors at the lexical and syntactic levels.

All automatic predictors of speech fluency derived from FBDS segments were strongly associated with subjective ratings. The higher the subjective ratings of speech fluency, the higher the rate of pseudo-syllables and speech ratio, and the lower the rate of silent breaks and the regularity of pseudo-syllable length. These relationships were expected, as speech rate is a known contributor of speech fluency, and silent breaks and filled pauses (which increase the variability in syllable length) are classic disfluency markers used to characterise the fluency of PWA (Faroqi-Shah et al., 2020; Gordon & Clough, 2022; Jacks & Haley, 2015; Wang et al., 2013).

When combined together into multivariate regression models, the automatic measures could achieve accurate predictions of reference speech fluency ratings, especially when taking into account several sentences per participant. As in the study of Fontan et al. (2022) who used the same predictors for measuring speech fluency in L1 Korean children, no difference was observed across regression models, and the two most important predictors were the speech ratio and the rate of pseudo-syllables. In the current study, the predictions are however slightly less accurate than in the study of Fontan et al. (2022) where the MLR achieved at the sentence level a RMSE of 0.35, against 0.52 in the present study. As both studies used relatively small samples (65 recordings in the case of Fontan et al., 2022), this small difference might be partially due to differences in the distribution of the reference fluency scores. An additional explanation is that in the present study, and contrary to the data collected by Fontan et al. (2022), a significant part of the recordings (15%) contained word repetitions, with up to seven words repeated in a single sentence. Yet, the algorithms developed by Fontan et al. (2022) do not account for such disfluencies. When syllables are repeated in the speech signal, they are taken into account by the algorithms in exactly the same way the other syllables are for the calculation of speech rate and speech ratio. However, contrary to "productive", non-repeated syllables that may have a positive impact on speech rate and speech ratio — and therefore on speech fluency, repeated syllables are disfluencies that affect perceived speech fluency in a negative way (Faroqi-Shah et al., 2020; Gordon & Clough, 2022). As this effect was not captured by any of the predictors, the accuracy of the models might not be optimal.

In order to take into account repetitions in read-speech data, a simple yet potentially effective solution would be to compute the difference between the number of pseudo-syllables that are automatically identified by the algorithm and the number of syllables that should theoretically have been pronounced according to the sentence script. To explore the relevance of this additional predictor, an MLR was computed over the whole 95 observations, and using either the three predictors retained by the stepwise selection (speech ratio, pseudo-syllable rate and standard deviation of pseudo-syllable duration; model 1 in Table 6) or the same three predictors plus the difference between the number of pseudo-syllables that were automatically detected and the expected number of syllables based on the sentence read by the participants (model 2 in Table 6). As can be observed, the new predictor significantly improves the coefficient of determination, which increases from 0.78 to 0.83. The RMSE also decreases from 0.53 to 0.47 (and down to 0.31 if aggregating the data for each participant). Figure 4 shows the scatterplots achieved when using the model 2 for predicting speech fluency ratings at the sentence or participant level.

Table 6. Results of multiple linear regressions using an estimate of the number of unexpected pseudo-syllables (model 2) or not (model 1).

| Model | Predictors | RMSE | $R^2$ change | p-value |
|---|---|---|---|---|
| 1 | Speech ratio, pseudo-syllable rate, standard deviation of pseudo-syllable duration | 0.53 | 0.78 | < 0.001 |
| 2 | All predictors used in model 1 + difference between the number of pseudo-syllables and the expected number of syllables | 0.47 | 0.83 | 0.005 |

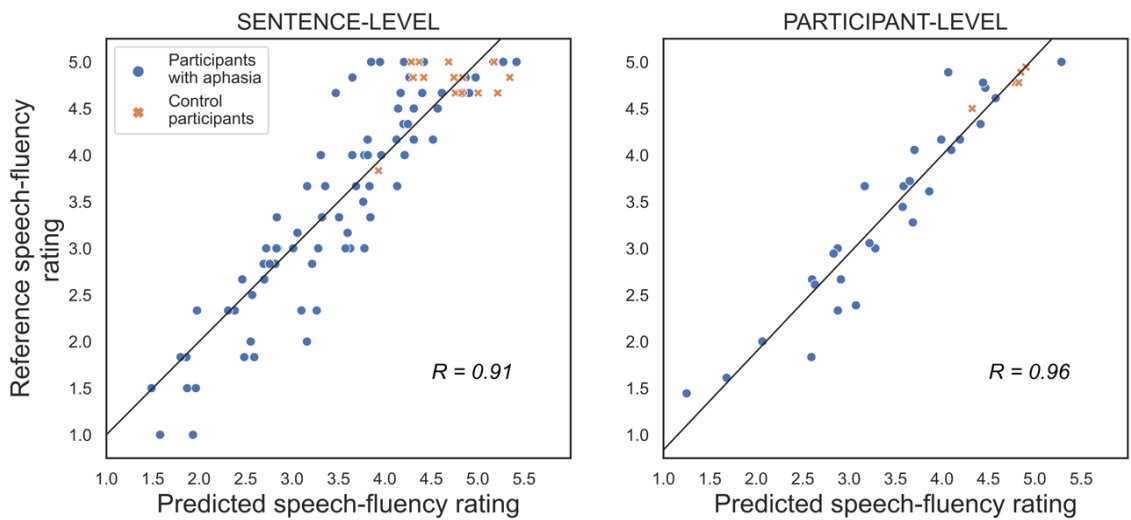

Figure 4. Scatterplots relating reference speech-fluency ratings to ratings predicted using an MLR, after adding the estimated number of syllable repetitions as a predictor, and associated regression lines. The left panel shows data at sentence level (95 data points), while the right panel shows data averaged for each participant (34 data points).

The study also showed that taking into account several sentences per participant could increase the accuracy of the predictions, as was already observed by Fontan et al. (2018; see Figure 2). When taking into account only three sentences (or less for the five participants who could not read all three sentences), the RMSE decreased from 0.52 to 0.4. Future research work should investigate if taking into account a higher number of sentences per participant can further increase the accuracy of the predictions. However, as any increase of the number of sentences would be at the expense of a reduced feasibility of the task in a clinical context, it should be determined up to which point this increase in accuracy is desirable.

All in all, the present study demonstrates that the FBDS algorithm, and the automatic schemes developed to cluster FBDS segments into pseudo-syllables and silent breaks, can be successfully used to predict the speech fluency of PWA, as assessed by trained SLPs and using a read-speech task. The results are all the more encouraging as speech was not recorded in acoustically-controlled, laboratory conditions but rather in challenging conditions, with participants using the built-in microphones of their own PCs during videoconferencing sessions from their home. This suggests that the algorithms used in this study are robust and could thus be used in clinical conditions (e.g., at the hospital or at the SLP clinic) using standard recording devices. The system could be used to rapidly assess, at a very low cost for the researchers or clinicians, the fluency of PWA in a task such as the reading task of the Boston Diagnostic Aphasia Examination (Goodglass et al., 2001). Such measures could be part of the multidimensional fluency profiles proposed by Clough and Gordon (Clough and Gordon, 2020; Gordon and Clough, 2022), by providing estimates for the "speech production" level, as a complement to higher-level metrics.

In the longer term, the relevance of these algorithms for the assessment of speech fluency of PWA during spontaneous interactions (e.g., during conversational dyads such as those recorded for the AADI project; Sahraoui et al., 2022) should be studied, as this kind of oral production has a better external validity than read-aloud tasks. However, using automatic signal segmentation techniques to predict speech fluency is far more complex for spontaneous speech, as in this case perceived fluency tends to be affected by nontemporal aspects of speech production such as grammatical and lexical errors (Cucchiarini et al., 2000; 2002). Another obvious difficulty with spontaneous speech is that, contrary to read speech, no reference script is available. To quantify the number of repetitions in spontaneous speech (i.e., to calculate the additional predictor aforementioned), advanced signal-processing algorithms should therefore be used, as those developed for the automatic detection of syllable repetitions in stuttered speech (Chee et al., 2009; Ramteke et al., 2016; Sahidullah et al., 2023).

Finally, due to the limited number of PWA in this study, as well as to their imbalance in terms of clinical profiles, the analysis of the relationships between automatic fluency scores and clinical variables was deemed beyond the objectives of the current proof-of-concept. Future research work, using a larger sample of PWA, is therefore warranted to investigate if and how automatic scores (overall fluency scores and associated predictors) relate to clinical indicators that may impact speech fluency, such as aphasia type, time post-onset, and data related to speech rehabilitation programmes.

**Acknowledgments**

The authors express their deep gratitude to all the participants who accepted being recorded, and to the three SLPs who participated in the fluency rating task. They also warmly thank Prof. Jean-Luc Nespoulous, Dr. Sébastien Déjean, and Dr. Saïd Jmel for their helpful advices. The study was funded by the European Regional Development Fund (ERDF), within the framework of the research project "Aphasie et Discours en Interaction (AADI) [Aphasia And Discourse in Interaction (AADI)]" (funding number: 2019-A03105-52). SMF also acknowledges support of Ramón y Cajal (RYC2020-028927-1), Ministerio de Ciencia e Innovación, Spain.

**Declaration of interest statement**

This study is part of the development, by Archean LABS (host institution of LF), of a software intended for speech-language therapists.

# Appendix A - Background information on the participants

## A.1 Participants with aphasia

| Participant | Sex | Age (years) | Education since 1st grade (years) | Aphasia type | Time post-onset (years) |
|---|---|---|---|---|---|
| A01 | Male | 63.4 | 24 | Broca | 25 |
| A02 | Male | 48.8 | 23 | Broca | 2.2 |
| A03 | Male | 71.1 | 15 | Wernicke | 10.3 |
| A04 | Male | 62.6 | 15 | Broca | 6.1 |
| A05 | Male | 54.3 | 22 | Broca | 17.1 |
| A06 | Male | 42.5 | 14 | Wernicke | 0.7 |
| A07 | Male | 67.4 | 23 | Broca | 21.4 |
| A08 | Female | 70.3 | 8 | Broca | 6.9 |
| A09 | Male | 60.7 | 10 | Wernicke | 5.5 |
| A10 | Male | 49.7 | 13 | Broca | 25 |
| A11 | Male | 49.4 | 16 | TCMi[a] | 2.8 |
| A12 | Male | 69 | 8 | Broca | 6.1 |
| A13 | Male | 61.4 | 11 | Broca | 7.5 |
| A14 | Female | 56.0 | 15 | Broca | 21 |
| A15 | Male | 55.9 | 15 | Mixed | 3.8 |
| A16 | Male | 27.1 | 16 | Wernicke | 1.9 |
| A17 | Female | 70.1 | 13 | Broca | 2.4 |
| A18 | Male | 67.4 | 10 | Broca | 15.1 |
| A19 | Male | 59.0 | 10 | Broca | 21.2 |
| A20 | Male | 74.5 | 12 | Broca | 24.5 |
| A21 | Female | 72.5 | 17 | Broca | 20.2 |
| A22 | Male | 68.1 | 12 | PPA[b] | 4 |
| A23 | Male | 60.4 | 24 | Broca | 19.9 |
| A24 | Female | 33.8 | 16 | PPA[b] | 2.8 |
| A25 | Male | 57.0 | 10 | N/A | 7.2 |
| A26 | Male | 70.0 | 11 | Broca | 24.4 |
| A27 | Female | 64.2 | 11 | Wernicke | 5.4 |
| A28 | Female | 69.7 | 20 | Broca | 9.4 |
| A29 | Female | 73.5 | 14 | PPA[b] | 1.6 |

[a] Transcortical mixed
[b] Primary progressive aphasia

## A.2 Control participants

| Participant | Sex | Age (years) | Education since 1st grade (years) |
|---|---|---|---|
| C01 | Female | 33.2 | 19 |
| C02 | Female | 74.8 | 12 |
| C03 | Female | 23 | 17 |
| C04 | Male | 73.2 | 20 |
| C05 | Female | 61.1 | 13 |